\documentclass{article}

\usepackage[english]{babel}

\usepackage[letterpaper,top=2cm,bottom=2cm,left=3cm,right=3cm,marginparwidth=1.75cm]{geometry}

\usepackage{amsmath}
\usepackage{graphicx}
\usepackage[colorlinks=true, allcolors=blue]{hyperref}

\title{Private Federated Learning In Real World Application -- A Case Study}
\author{An Ji, Bortik Bandyopadhyay, Congzheng Song\\
Natarajan Krishnaswami, Prabal Vashisht, Rigel Smiroldo, Isabel Litton\\
Sayantan Mahinder, Mona Chitnis, Andrew W Hill\\
Apple}

\begin{document}
\maketitle

\begin{abstract}
This paper presents an implementation of machine learning model training using private federated learning (PFL) on edge devices. We introduce a novel framework that uses PFL to address the challenge of training a model using users' private data. The framework ensures that user data remain on individual devices, with only essential model updates transmitted to a central server for aggregation with privacy guarantees. We detail the architecture of our app selection model, which incorporates a neural network with attention mechanisms and ambiguity handling through uncertainty management. Experiments conducted through off-line simulations and on device training demonstrate the feasibility of our approach in real-world scenarios. Our results show the potential of PFL to improve the accuracy of an app selection model by adapting to changes in user behavior over time, while adhering to privacy standards. The insights gained from this study are important for industries looking to implement PFL, offering a robust strategy for training a predictive model directly on edge devices while ensuring user data privacy. \end{abstract}

\section{Introduction}

Behavior modeling and app prediction are essential for creating intelligent virtual assistants and personalized user experiences. These techniques\cite{Tongaonkar2013UnderstandingMA}, \cite{Wang2020AHA}, \cite{Yu2017SmartphoneAU} analyze user interactions to forecast behaviors, such as app usage or feature preferences, using machine learning algorithms.
With the rise of mobile usage, accurately predicting user interactions is key to improving engagement and personalizing content. However, this process often requires extensive data collection, raising privacy concerns.
Private Federated Learning PFL \cite{Fu2024DifferentiallyPF} addresses these concerns by enabling decentralized training of machine learning models, keeping data on the user's device while sharing only model updates with privacy guarantees. Wu et al. \cite{Wu2021HierarchicalPF} propose a hierarchical personalized federated learning (HPFL) framework that enhances user modeling by balancing global patterns with local personalization, all while maintaining data privacy.

In this work, we apply the PFL framework to train a predictive model for app selection, ensuring user privacy by keeping data local on individual devices. PFL allows each device to compute model updates based on its own data, which are then aggregated by a central server to refine a global model without ever accessing raw data. This approach not only preserves privacy but also enables better personalization through the HPFL framework, which tailors models to individual users while maintaining the benefits of a collective learning process. Section 2 will present the modeling and PFL details. Followed by offline simulation experiments and on device training set up in Sections 3 and 4. In the end, we will share our learnings and conclusions for using PFL techniques to train industrial applications. 
\section{Preliminaries} 

\subsection{App selection overview}
The goal of App Selection is to provide third-party applications with equal access to the virtual assistant ecosystem by enabling them to respond to user requests within the domains they support, without the user needing to specify the app by name. This allows for a more seamless and user-friendly experience.

For example, when a user makes a request to play music, App Selection uses the user’s habits to determine the most suitable app. If a particular app is frequently used for similar requests, the model will prioritize launching that app, even if it wasn’t specifically mentioned.

By using Private Federated Learning (PFL), we ensure that no user's app usage data from any individual device is collected, while still improving the overall algorithm across all devices. This process allows us to continuously refine and enhance the model's accuracy by running it on devices, while preserving user privacy. Updated models are then pushed back out to all users for a better overall experience. 

\subsection{App selection model}
\label{sec:model_desc}
Figure 1 shows high-level app selection model architecture:

\begin{enumerate}
    \item \textbf{Entity-type-specific Data Preparation:} This  focuses on preparing and processing data within the model rather than in the client code, allowing for dynamic updates and flexibility in data handling.
    \item \textbf{Cross-entity Feature Engineering:} At the core of the model is the cross-entity Feature Engineering module, utilizing a multi-headed attention mechanism. This module evaluates the interplay between different apps and associated signals to ascertain the most contextually relevant factors for prediction.
    \item \textbf{Epistemic Uncertainty Handling:} This component assesses the model's predictive confidence. It evaluates whether the model is operating within its competent predictive boundaries or merely speculating. If the predictive confidence is low, it suggests that none of the app candidates are suitable, prompting a user interaction for clarification.
    \item \textbf{Aleatoric Uncertainty Handling:} In contrast, this stage manages uncertainty stemming from having multiple similar app candidates. It distinguishes situations where multiple apps appear equally viable, suggesting a user prompt for clarification, from those where one app significantly outweighs others, thus warranting a direct execution.
    \item \textbf{Action Selection:} The final component synthesizes the outcomes of the uncertainty assessments to recommend definitive actions for each app candidate. Depending on the context, this could range from directly launching an app to providing a list for user disambiguation. This process is intricately designed in collaboration with human interface design principles to ensure intuitive user interactions.

\end{enumerate}

\begin{figure}
\centering
\includegraphics[width=0.8\textwidth]{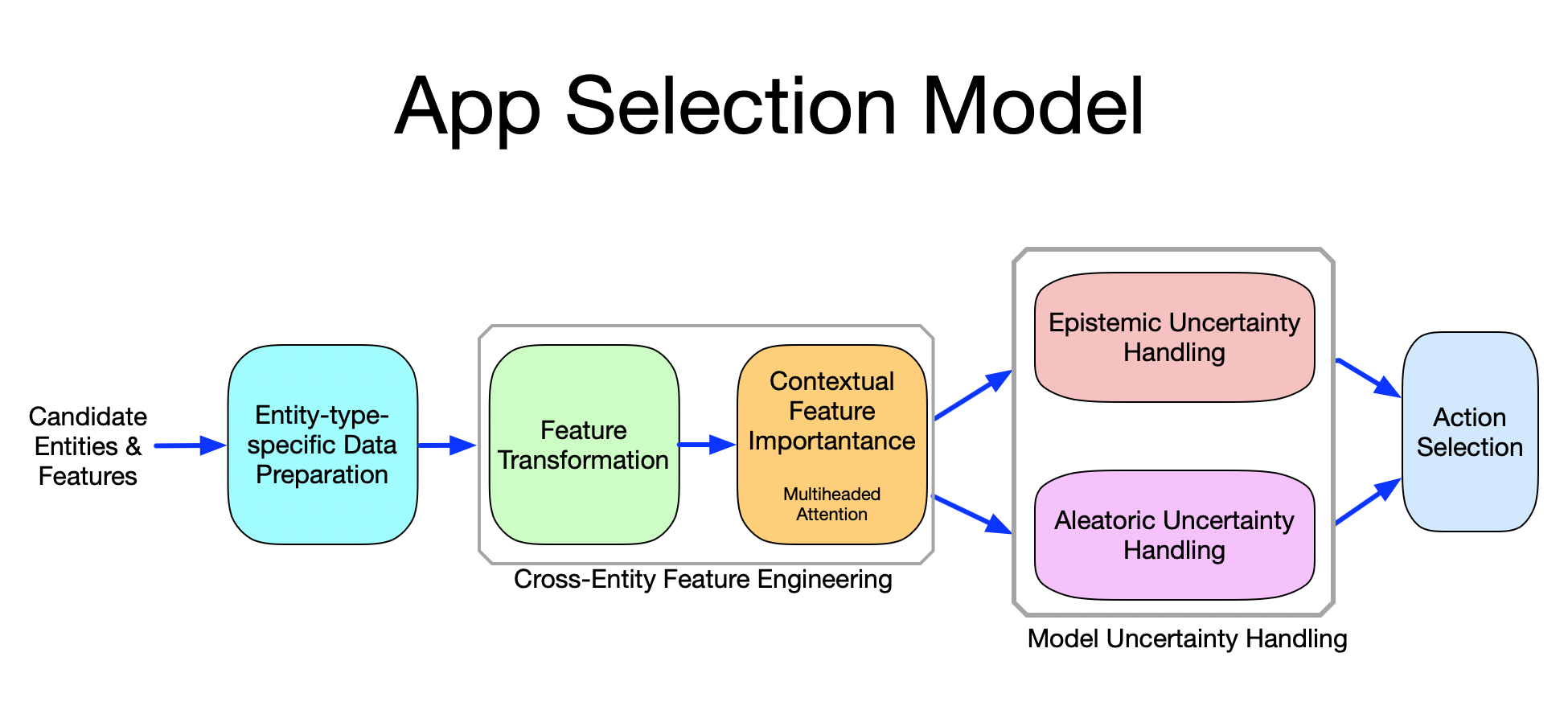}
\caption{\label{Figure 1: App Selection}Model architecture}
\end{figure}

\subsection{Evaluation metrics}
\label{sec:eval_metrics}
We used two sets of metrics to evaluate the performance of our model. To monitor model training for the Deep Neural Network (DNN), we used mean square error (MSE) loss and used accuracy to measure offline metrics. These offline metrics measure the percentage of interactions in which the direct execution of the model aligns with the user selection.

To assess the model's performance in user-facing scenarios, we introduced correct direct execution rate and disambiguation rate, referred to as our online metrics.
\begin{itemize}
    \item \textbf{Correct Direct Execution Rate (CDER):} This metric represents the percentage of interactions where the model's direct execution correctly corresponds to the user's intended action.
    \item \textbf{Disambiguation Rate:} This metric quantifies the percentage of interactions where the model would naturally present a disambiguation prompt to the user.
\end{itemize}

\subsection{Private federated learning}

Federated learning (FL) \cite{McMahan2017LearningDP} is a distributed machine learning approach that enables multiple participants, often devices or data centers, to train a model while keeping all training data locally. As articulated by \cite{McMahan2017LearningDP}, FL allows devices to contribute in the creation of a shared model by computing updates locally on their own data and then sending these updates to a central server to be averaged. This technique helps to protect user data and reduces the need for data transfer, which can expose data to additional risks during transmission \cite{Konecn2016FederatedOD}. FL is particularly advantageous where data cannot leave its original location due to privacy, security, or logistical constraints, making it a popular choice for industries.

However, FL on its own does not provide privacy because one can learn from the model updates about the training data. Incorporating Differential Privacy (DP) \cite{McMahan2016CommunicationEfficientLO} into FL frameworks improves privacy assurances by adding stochastic noise to model updates, ensuring that the contributions of individual devices are not discernible in the aggregated data received by the central server. \cite{Dwork2006CalibratingNT} introduced differential privacy as a method for providing strong privacy guarantees, defining it in terms of limiting the risk of identification from the output of database queries, irrespective of any auxiliary information that may be available. By integrating differential privacy into federated learning, as demonstrated by \cite{Abadi2016DeepLW} in their implementation of a differentially private stochastic gradient descent algorithm, it becomes significantly more challenging to infer information about any individual participant's data from the shared model. This dual approach of FL and DP is pivotal in scenarios where privacy is paramount, such as in personalized medicine or financial services, but it introduces challenges in balancing privacy, accuracy, and computational efficiency, requiring ongoing optimization and innovation in algorithmic strategies \cite{Geyer2017DifferentiallyPF}.
Differential Privacy (DP) \cite{Dwork2014TheAF} inherently impacts model accuracy by introducing noise to protect individual data points during the learning process. The noise added to either the data itself or during the aggregation of model updates ensures that the contributions of individual participants cannot be discerned, thereby protecting privacy. However, this addition of noise can degrade the model's performance because it obscures the underlying data patterns that the model aims to learn. The degree of noise correlates directly with the level of privacy guaranteed; higher privacy typically results in higher noise levels, which can further reduce accuracy. To mitigate the adverse effects of DP on model accuracy, researchers have developed several strategies. One effective approach is to carefully allocate the privacy budget across multiple iterations, balancing the trade-off between the amount of noise added and the need for accuracy. The way this budget is distributed can significantly impact the model's performance while still ensuring privacy. Techniques such as adaptive clipping and privacy budget allocation can optimize the use of the budget throughout training \cite{Xia2022DifferentiallyPL}, \cite{Hong2021DynamicPB}, concentrating privacy expenditure during the phases when it is most beneficial for learning. Additionally, improving the efficiency of the underlying learning algorithms, such as using more suitable machine learning models and optimization algorithms that are inherently more robust to noise, can also help offset the impact of differential privacy on accuracy.

\section{Feasibility Study using Offline PFL simulations}

Before we can implement Private Federated Learning (PFL) on devices, it is essential to begin the process with comprehensive offline simulations. We use Apple's public framework \texttt{pfl-research}\footnote{https://github.com/apple/pfl-research}, which provides a platform for running offline PFL simulations with any available central data, without requiring actual user devices. This simulation environment replicates the behavior of PFL during real-time training with user devices.  Such offline simulations offer critical insights into the potential efficacy of PFL for the intended application, allowing practitioners to assess the balance between privacy protection and utility even before PFL deployments. Moreover, the use of simulations facilitates extensive hyperparameter tuning on a large scale. This capability is essential for determining the optimal settings for both learning and privacy-specific hyperparameters, ensuring that the model performs effectively while adhering to required privacy constraints. This preparatory step is pivotal for ensuring that the adoption of PFL is both practical and aligned with the specific needs and constraints of the application at hand.

The App Selection model described in Section~\ref{sec:model_desc} consists of a Deep Neural Network model, whose weights can be learned or updated using Private Federated Learning~\cite{reddi2020adaptive} . We have evaluated offline PFL simulations on the DNN part of the model in two different ways.:
\begin{itemize}
    \item \textbf{Training from scratch} : In this setting, all weights of the neural network are initialized randomly and then learned as part of the back-propagation step. Thus the model is trained from scratch, which makes this a more complex and time consuming task. It is a relatively higher data and compute hungry setup, but can be used to support breaking changes in the model architecture or adding new features to the modeling task. 
    \item \textbf{Fine-tuning from an existing checkpoint} : In this setting, we start from an already trained model checkpoint (i.e., the weights have been learned previously) and then only a subset of weights of the neural network are unfrozen i.e., updated as part of back-propagation step, while the remaining weights are frozen. This setting is particularly useful when we already have a fixed architecture, fixed set of features \& a solid baseline model to which we will simply add data to continuously update the model to adapt to distribution shift (need citation).
\end{itemize}

Unless otherwise mentioned, we have used AdamW~\cite{loshchilov2017decoupled} as the server side optimizer and SGD as the client side optimizer for PFL simulations using \texttt{pfl-research}. 
The PFL model’s performance is evaluated using the accuracy metric defined in Section~\ref{sec:eval_metrics} computed on the fixed validation set.
Note that the training and evaluation data required for each of the two above setups are different, and hence are elaborated in the subsequent sections.

\subsection{Training from scratch}
\label{sec:scratch_train}

In this setup, we want to simulate the PFL-based training from scratch with Differential Privacy (DP) enabled. 
We start the training with a randomly initialized set of model parameters and train the entire model from scratch using a relatively large offline data set consisting of $\sim$788K data points. 
As the baseline, we have trained the same architecture with the entire training data from scratch without any PFL, which we refer to as \textit{Cymba} for simplicity and ease of reading.
We have a dedicated validation set consisting of $\sim$178K data points, which we use to compare the performance of the PFL trained model against the non-PFL baseline (i.e., Cymba).

Unless otherwise specified, we have used Gaussian Moments Accountant (Needs citation) implemented in \texttt{pfl-research} as the central privacy mechanism, with parameters as : $\epsilon = 2.0$, $\delta = 1e-6$, and $\text{Clipping Bound} = 0.1$. 
Additionally, for simplicity of offline simulation, we have fixed the \textit{mean data points per user} $= 1$ and \textit{local epochs} $= 3$, after some initial hyper-parameter exploration.
In all our experiments, we have set a higher Local Learning rate (LLR) for the on-device SGD step due to sparsity of training data per device, and a lower Central Learning rate (CLR) for the server side Adam optimizer step. This configuration has generally helped us achieve a good distributed training set up using PFL, while reducing the risk of training divergence due to sub-optimal hyper-parameter choice.

We design the offline simulations in such a way that the insights from these simulations can help plan the duration of on device training too.
Note that for the app selection model, there are two very important parameters that determine the overall time and resources needed for on device training viz. the \textit{number of devices per Central iteration} and the \textit{number of Central iterations}. 
We chose two sets of values to represent the High and Low resource Budget settings to determine the performance bounds of the PFL training :
\begin{itemize}
    \item \textbf{High Resource Budget} : In this setting, we choose 10K devices per central iteration, and 500 number of central iterations. Thus we take approximately $(10\text{K} * 500)/788\text{K} = 6.3$ passes on the entire training data, which is smaller than the total number of epochs used to train the non-PFL Cymba baseline.
    \item \textbf{Low Resource Budget} : In this setting, we choose 1K devices per central iteration, and number of central iterations = 500. Thus we take approximately $(1\text{K} * 500)/788{\text{K}} = 0.63$ passes on the entire training data, which is smaller than the total number of epochs used to train the non-PFL Cymba baseline.
\end{itemize}
We observe that in the high resource budget setting, the PFL trained model is almost similar in performance compared to Cymba, with a very minor regression that is acceptable, given that Differential Privacy will have a negative impact on the model's learning capacity.
As expected, the low resource budget setting appears to have significantly more regression compared to Cymba, likely because of the significantly smaller number of passes on the training data, coupled with the impact of Differential Privacy. 
Note that in this setting, it is difficult to achieve an improvement over an existing non-PFL trained baseline using the exact same training data.
However, given the PFL trained model's performance under the High Resource Budget setting, we consider that we may be successful in training a PFL model from scratch during on device training, which may have a competitive performance (i.e., within acceptable limits of performance regression) compared to the non-PFL trained baseline.

\begin{table}
\centering
\begin{tabular}{l|l|l|r}
Model & CLR & LLR & Accuracy\\\hline
Cymba & - & - & 0.856\\
PFL with Low Resource Budget & 0.001 & 0.01 & 0.83 \\
PFL with High Resource Budget & 0.0005 & 0.01 & 0.852
\end{tabular}
\caption{Performance of Training from Scratch with v/s without PFL}
\label{tab:scratch_train} 
\end{table}

Additionally, as part of offline simulations, we did multiple rounds of hyper-parameter tuning by varying the number of devices per central iteration, number of Central iterations, Central learning rate, Local number of epochs, Local learning rate and Privacy Clipping Bound before selecting the above configuration.
Some key observations are presented below:
\begin{itemize}
    \item We fixed the Local Learning Rate to a relatively higher value (0.01) as it is a bit difficult to modify/control this parameter during on device training. However, we varied the Central Learning Rate and observed that a value between [0.1, 0.9] often causes the training to diverge, while a value of 0.0005 tend to yield consistently good results on the offline data. Additionally, we tested various learning rate scheduling strategies (like Polynomial, Cyclic etc.) for the Central Learning rate, but did not observe any major gains over a fixed Central Learning rate.
    \item We noticed that the Local Number of Epochs  $\leq3$ tends to give better results. Increasing the number of epochs any further causes training divergence, even with small values of Local Learning Rate.
    \item We varied the number of Central Iterations from 500 to 25,000 but the benefits in terms of gains gradually decreases. Hence we fix the number of Central iterations to 500 (which is much smaller than 2K), as it will reduce the on device training time significantly when compared to the time taken for 2K iterations, given that the difference in performance is very small.
    \item Assuming one data sample per device, we varied the number of Devices between 1K and 150K. The general observation is that 5K to 10K devices tended to yield good results. If we choose 5K for on device training, then it will speed up the training time as well as reduce the network communication load with the backend PFL server.
\end{itemize}

\subsection{Fine-tuning from an existing checkpoint}
\label{sec:finetune_chkp}

In this setup, we start from an already trained model checkpoint (i.e., the weights have been learned previously using a different data set) and then train a subset of weights of the neural network using random sampled data. 
In this case, Cymba is the base model from which we start the training, and use a random sampled batch of $\sim$814K data points as training data, and another random sampled batch of $\sim$176K data points as Validation set, for performance comparison.
We test two variations to establish the performance bounds: a)~training all layers of the pre-existing model and b)~training only the top layer of the pre-existing model, while freezing the remaining layers.
We explore these 2 variants with the goal to reduce the number of PFL trained parameters in the model, as it is easier to train a model with fewer parameters since it reduces the on-device training complexity, lowers the network communication cost between the device and server as well as helps with Privacy.

\begin{figure}
\centering
\includegraphics[width=\textwidth]{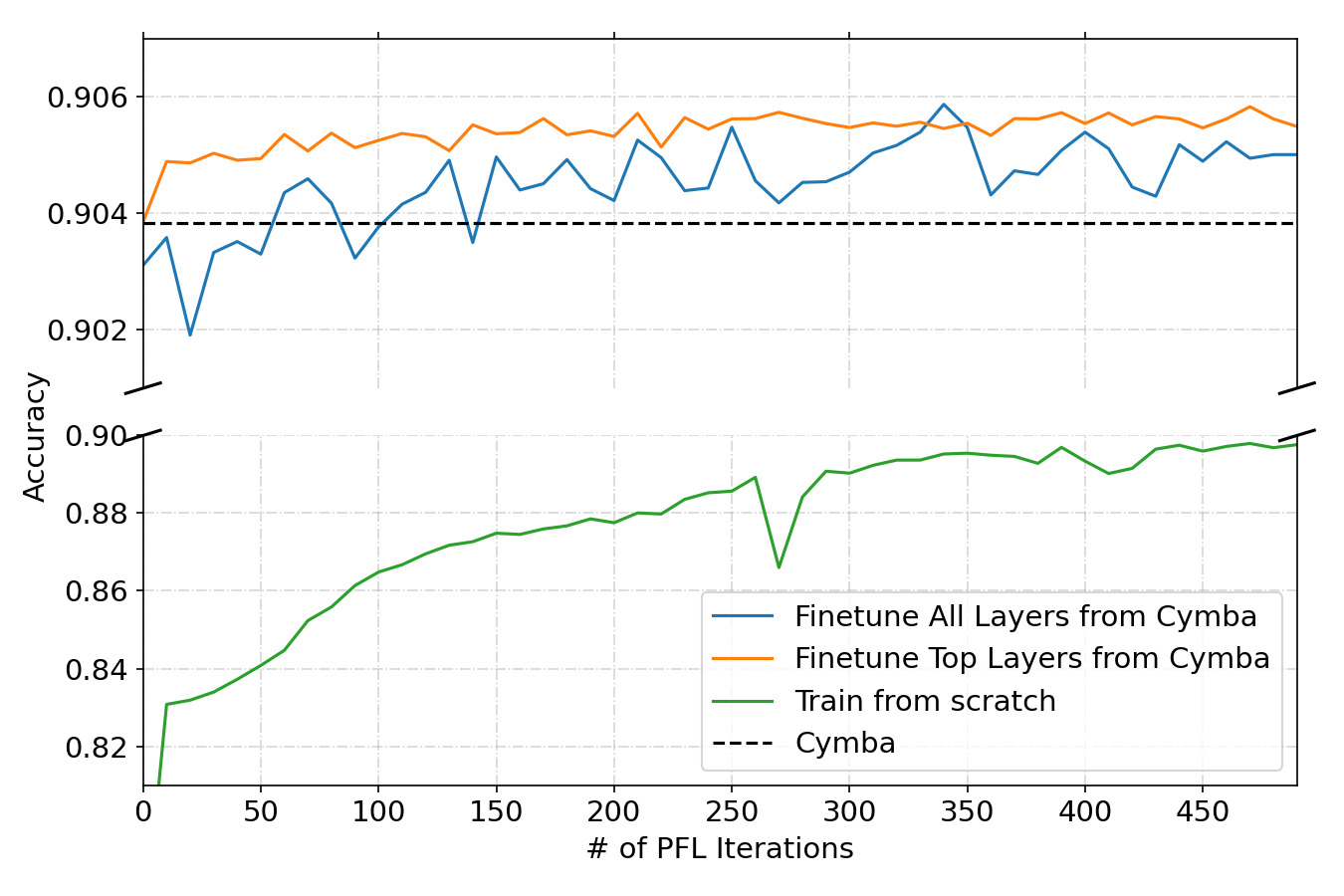}
\caption{Fine-tuning Cymba from an existing checkpoint}
\label{fig:fine_tune_cymba}
\end{figure}

The Accuracy metric of the PFL fine-tuned models are presented in the Y-axis in Figure~\ref{fig:fine_tune_cymba}, while the X-axis refers to the number of PFL Central Iterations for each of the different training config.
The performance of the Cymba model on this random sampled Validation set is also reported for tracking purposes.
Unless otherwise specified, we have used Gaussian Moments Accountant (Needs citation) implemented in PFL-Research(needs citation) as the central privacy mechanism, with parameters as : $Epsilon = 2.0$, $Delta = 1e-6$ and $Clipping Bound = 0.1$. 
For fine-tuning from existing checkpoint, we have set the \textit{mean data points per user} $= 1$, \textit{Local Number of Epochs} $= 1$, \textit{number of devices per Central iteration} $= 5000$ and the \textit{number of Central iterations} $= 500$. 
For fair comparison, we have trained a dedicated model from scratch (labeled as \textit{Train from scratch} in Figure~\ref{fig:fine_tune_cymba}) using the random sampled $\sim$814K training data points, while reducing the \textit{number of devices per Central iteration} $= 5000$.
We have done a hyper-parameter search for the learning rate and chose the best configuration for each setting when reporting the above metrics.

In Figure~\ref{fig:fine_tune_cymba}, we observe that fine-tuning with freshly random sampled training data improves the performance of the PFL trained models on Validation set, compared to the static Cymba baseline. 
This indicates that the app selection model can adjust to the user's behavioral shift over time (as represented within the random sampled Training and Validation sets) using PFL, which is very useful for continuous model maintenance/upgrade activity.
Note that fine-tuning only the top layer of Cymba appears to be performing better than fine-tuning all layers,which reduces the amount of communication bandwidth spent to transfer the PFL weights from device to the servers, thereby making PFL less network bandwidth-intensive operation. Also fewer learnable parameters is better for DP thereby making this training paradigm a more suitable one for this use-case.
Finally, using PFL to train a model from scratch using random sampled data still under-performs all the settings, but the gap in performance is smaller if hyper-parameter tuning can be done appropriately.

\begin{figure}
\centering
\includegraphics[width=\textwidth]{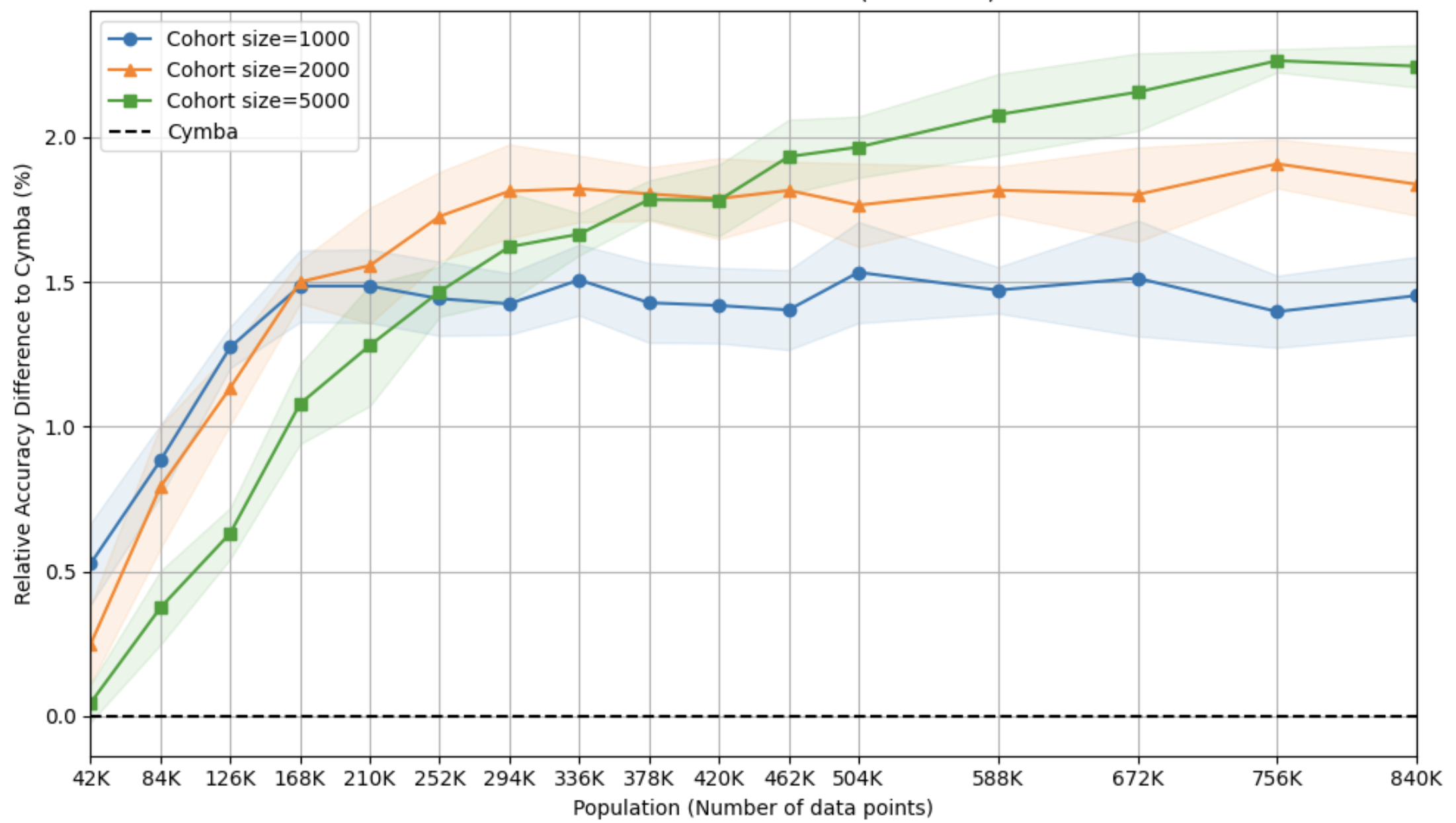}
\caption{Using simulations to predict the performance of the fine-tuning of Top Layers of the Cymba model as a function of Number of devices (cohort size) and corresponding total number of data points.}
\label{fig:data_req_for_tuning}
\end{figure}

In Figure~\ref{fig:data_req_for_tuning}, we plot the performance of fine-tuning the Top Layers of the Cymba model by varying the \textit{number of devices per Central iteration} as (1K, 2K, 5K). Assuming \textit{mean data points per user = 1}, in the X-axis we plot the number of data points that have been used for fine-tuning the top layers of Cymba for the corresponding Cohort size. The Y-axis shows the relative difference in Accuracy of the corresponding model checkpoint compared to the performance of the Cymba model on a fixed Validation set. This plot gives us an approximate insight into the amount of data points required to obtain a certain percentage of relative accuracy improvement, which will help us to select the corresponding parameters and the duration of the on device training. We observed that with a Cohort size of 2K while we achieve an improvement with fewer data points, model performance plateaus quickly, even though the performance can be further improved with more data points using a Cohort size of 5K. For example, with 504K data points, we can achieve close to 2\% relative improvement in Accuracy with a Cohort size of 5K, while the relative improvement is lesser with a Cohort size of 2K. This shows the importance of selecting the Cohort size appropriately in conjunction with the duration of the on device training (i.e., total number of data points to use for fine-tuning) to achieve best improvement.

%\subsection{Ablation study}
%\textbf{Training data retention period estimation experiments}

\section{PFL on device training}

\subsection{On device training data}

Training data are generated during inference time through user's explicit feedback.  These data are stored on device. Each training record includes feature values, ground truth labels and metadata. On device data storage system provides a mechanism whereby a particular task can filter out records which satisfy certain matching criteria specified by PFL server. For example, it is possible to match on device OS versions, or target a specific set of data produce by a specific on-device asset through this mechanism.  

\subsection{Federate statistics}

We use Federated Statistics, \cite{CorriganGibbs2017PrioPR}, which is Apple’s end-to-end platform for learning histogram queries from sensitive data on-device, to run histogram queries from on-device data to gain training data insights, such as how much data are available to participate in the PFL training. Before we launch PFL training, as part of the feasibility study, using the FedStats query, we found that ABSOLUTE NUMBER of devices have at least 1 valid sample to participate in PFL training thereby satisfying our data requirements. This shows us that we can complete PFL training iterations with reasonable latency and achieve model convergence. 

\subsection{On device plug-in design}

To enable real devices to process local data and contribute to a Personalized Federated Learning (PFL) task, an on-device plugin was developed. The primary objectives of the plugin are to process local data stored on the device using parameters defined in the PFL task description and attachments, compute a model update or generate training statistics and metrics, and then send these results to a central server for aggregation.
The plugin also includes an on-device differential privacy component. It ensures user privacy and security by applying differential privacy (DP) techniques and encrypting the model updates or training statistics before transmission, protecting sensitive data throughout the process.

\begin{figure}
\centering
\includegraphics[width=\textwidth]{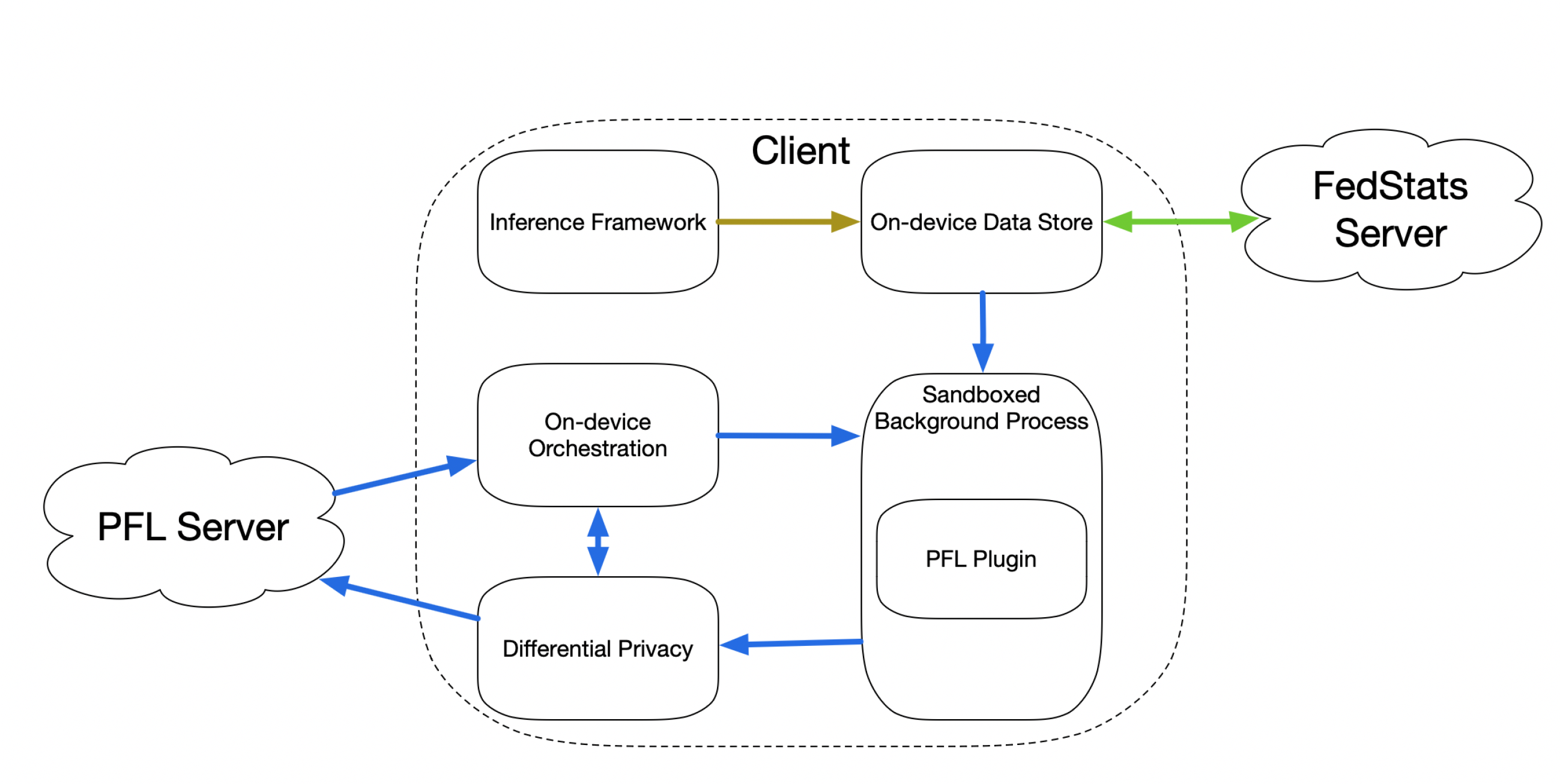}
\caption{PFL on device training workflow}
\label{fig 4}
\end{figure}

The training workflow shown in Figure~\ref{fig 4} involves several key components to ensure on-device data processing and training while maintaining privacy and security:

\begin{enumerate}
    \item \textbf{Inference Framework and On-device Data Store:} The inference framework collects training data and stores it in the On-device Data Store, which is essential for PFL tasks.
    \item \textbf{Data Utilization by FedStats Server:} The FedStats Server utilizes the data stored on the device to aggregate statistics or provide insights into data distribution without directly accessing raw data.
    \item \textbf{PFL Plugin and On-device Orchestration:} The PFL plugin processes local data using task descriptions and parameters provided by the On device orchestration, which communicates with the PFL Server to receive these descriptions and attachments. This runs in a secure and isolated environment as sandboxed process.
    \item \textbf{Differential Privacy and Encryption:} After processing the data and computing model updates or training statistics, these outputs are passed through the Differential Privacy component. It applies DP techniques to anonymize and secure the data, adding noise to protect user privacy. The aggregated and encrypted updates are then sent back to the PFL Server.
    \item \textbf{Model updates at PFL Server:} The PFL Server aggregates updates from multiple devices, contributing to overall model training without compromising individual data privacy.
\end{enumerate}

\subsection{On device evaluation}

To process local records, the plug-in will use the task sent from the server to devices, as well as model files and any additional files required by the plug-in. For example, in our case, app selection model can be trained on-device and the difference between the model parameter values before and after on-device training represent the result to be aggregated on a server for PFL training. 

We use custom-built tool to configure the necessary parameters for training a model or computing statistics on-device for a particular PFL task. Metrics will be computed at this stage in the plug-in, which will also be sent to a server for aggregation. When training ML models using PFL, metrics include training loss and training/evaluation accuracy. We also compute additional user facing metrics in Section 2.2. Metrics will be sent in the metadata field to the server from devices, along with encrypted results. 

\subsection{PFL training results}

We conducted several training cycles on different sizes of traffic. The results are summarized in Table~\ref{Result},

\begin{table}
\centering
\begin{tabular}{c|cc}
Model&CDER&Disambiguation Rate\\\hline
Baseline&89.18\%&1.99\%\\
PFL trained model&89.86\%&1.99\%\end{tabular}
\caption{\label{Result}Model Evaluation Results}
\end{table}
The PFL trained model showed about 0.6\% of absolute gain in CDER while keeping the same Disambiguation rate over our baseline model. The model’s gain is mainly due to users' change in behavior over time. The old (baseline) model trained on older server side data has drifted away from more recent data. The PFL model was trained on more recent data which captured this distribution change in user behavior.

We have also A/B tested this PFL trained model. Our A/B experiment was conducted for 2 weeks on about 15M devices. The PFL trained model achieved a 0.07\% gain in the top-line metric of system task completion rate and a 15.6\% decrease in the Disambiguation rate. This indicates our PFL trained model improved user experience by correctly predicting users' intended apps.

\section{Conclusion}
We demonstrated the potential of Private Federated Learning (PFL) in the domain of app selection with its capability to achieve accurate predictive modeling while protecting users' privacy. Our experiments, both in offline simulations and training environments, validate the PFL approach, showing that it can learn and adapt to the distribution shift of user preferences over time, while while guaranteeing privacy through model training on-device rather than collecting data and training at the server. A couple of learnings we would like to highlight: 

\begin{enumerate}
    \item \textbf{Unified Training Data Generation:} To get the most benefit from offline simulations, it is critical to standardize the process of generating training data both offline and on-device. This ensures the data is representative of real-world scenarios to avoid incorrect conclusions from simulations due to data biases.
    \item \textbf{Model Training Approaches:} When a pretrained model is available, opting to fine-tune from an existing checkpoint generally yields better results than starting the training process from scratch. Fine-tuning leverages the learned weights of the pretrained model, allowing for quicker convergence.
    \item \textbf{On-device Plugin Design:} Designing on-device plugins with minimal domain-specific code is beneficial. This approach facilitates easier generalization and adaptation of the plugin across different applications.
    \item \textbf{Pre-Training Data Verification:} Before initiating PFL on device training, it is crucial to verify the on-device training data. This verification process helps identify the amount of training data necessary to complete PFL training iterations with reasonable latency to achieve model convergence.
    \item \textbf{Crucial PFL Hyper-parameters:} Within the realm of PFL, certain hyper-parameters, particularly learning rate and cohort size, are critical, especially when fine-tuning models. Adjusting these parameters appropriately can significantly influence the effectiveness and efficiency of the federated learning process, impacting overall model performance and training speed.
\end{enumerate}

Future work will focus on optimizing the trade-offs between model accuracy and privacy, exploring more efficient ways to manage computational resources, and expanding the applicability of PFL to other areas of predictive modeling.

\bibliographystyle{abbrv}
\bibliography{main}

\end{document}